# Learning and Recognizing Human Action from Skeleton Movement with Deep Residual Neural Networks

**Huy-Hieu Pham**[†,††], **Louahdi Khoudour**[†], **Alain Crouzil**[††], **Pablo Zegers**[†††], **and Sergio A. Velastin**[††††]

[†] Centre d'Etudes et d'Expertise sur les Risques, l'Environnement, la Mobilité et l'Aménagement, France.
[††] Institut de Recherche en Informatique de Toulouse, Université Paul Sabatier, France.
[†††] Facultad de Ingeniería y Ciencias Aplicadas, Universidad de los Andes, Chile.
[††††] Department of Computer Science, University Carlos III Madrid, Spain.
{huy-hieu.pham;louahdi.khoudour}@cerema.fr, alain.crouzil@irit.fr,
pzegers@miuandes.cl, sergio@velastin.org

## Abstract

Automatic human action recognition is indispensable for almost artificial intelligent systems such as video surveillance, human-computer interfaces, video retrieval, etc. Despite a lot of progresses, recognizing actions in a unknown video is still a challenging task in computer vision. Recently, deep learning algorithms has proved its great potential in many vision-related recognition tasks. In this paper, we propose the use of Deep Residual Neural Networks (ResNets) to learn and recognize human action from skeleton data provided by Kinect sensor. Firstly, the body joint coordinates are transformed into 3D-arrays and saved in RGB images space. Five different deep learning models based on ResNet have been designed to extract image features and classify them into classes. Experiments are conducted on two public video datasets for human action recognition containing various challenges. The results show that our method achieves the state-of-the-art performance comparing with existing approaches.

## 1 Introduction

Human action recognition in video is an important research area in computer vision. Recognizing correctly actions in unknown video is a challenging task due to many factors such as occlusions, viewpoint, lighting and so on. Many approaches have mainly focused on recognizing actions from video sequences with RGB, depth, or combining these two data types (RGB-D). Currently, depth cameras such as Kinect is able to provide a powerful skeleton tracking algorithm [24] in realtime. Meanwhile, human actions can be represented by the movements of skeleton joints, thus using the skeleton data can distinguish many actions. Furthermore, the skeleton data representation has the advantage of lower dimensionality than RGB-D representations; this benefit makes action recognition models simpler and faster. For that reason, exploiting the 3D human joint positions from depth cameras for recognizing human action is a very effective research direction.

In recent years, deep learning based approaches achieved outstanding results in image recognition and classification. Among deep learning-based models for action recognition, Convolutional Neural Networks (ConvNets) were seen as the most important architecture. Many authors have proposed the use of ConvNets for solving problems related to human action recognition [25]. Most of these approaches learn action features from RGB-D image sequences and use learned features to classify actions. Although RGB-D images are very informative for action recognition, the computation complexity of the learning model will increase rapidly when the number of frames increases. Therefore, it makes models more complex and slower.

In this paper, we aim to take full advantages of 3D skeleton joints provided by Kinect sensor and the power of very deep ConvNet to build an end-to-end learning framework for human action recognition in video. Firstly, the skeleton data is collected by Kinect in frames. These skeleton sequences are then transformed into 3D-arrays and saved in RGB images in order to ensure they are accepted as the input of our ConvNet models. To learn and recognize actions, we propose the use of Deep Residual Network (ResNets) [12], a recent state-of-the-art ConvNet on image recognition and classification. Five different ResNet models have been designed and tested. Experimental results on two benchmark action datasets show that our leaning framework achieves the state-of-the-art accuracies in the same experimental condition. To the best of our knowledge, we are the first to use successfully ResNets for skeleton based action recognition in video.

The rest of the paper is organized as follows: Section 2 introduces the related works on using skeleton for action recognition. In section 3, we present our proposed models including a data transformation module, the principle of ResNet and five different ResNet architectures for recognizing actions. Datasets and their experimental protocols are described in Section 4 and results are shown in Section 5. Finally, conclusion is provided in section 6.

## 2 Related Work

Taking the advantages of depth camera technology, many solutions which use skeleton data have been proposed. Looking at the literature of human skeleton based action recognition, two main research directions can be found containing approaches based on local features and approaches based on Recurrent Neural Networks (RNNs).

*Approaches based on local features*: Human action is a spatio-temporal pattern, many researchers have proposed the use of temporal models for modeling human motions. Firstly, spatio-temporal features of an action are extracted. Then, the authors employ generative models such as Hidden Markov Model (HMM) [20, 32] or Conditional Random Field (CRF) [11] for modeling and recognizing the human action. Another technique called Fourier Temporal Pyramid (FTP) [29, 32] has been also used to describe the temporal structure of an action and predict its class. Although promising results have been achieved from these above approaches, there are some limitations which are very difficult to overcome. E.g., HMM based methods require preprocessing input data. Normally, the skeleton sequences need to be segmented and aligned. Meanwhile, FTP based approaches can only utilize limited contextual information of an action and cannot globally capture the temporal sequences of actions.

*Approaches based on RNNs*: Recurrent Neural Networks (RNNs) with Long Short-Term Memory Network (LSTM) [13] are able to model the contextual information of the temporal sequences as skeleton data. Thus, many authors have explored RNN-LSTMs for 3D human action recognition [5, 28]. Experiments provided that RNN-LSTMs outperform many previous works. However, the use of RNN can face to overfitting when the number of features is less in training phase. Moreover, except the study of Du *et al.* [5], all the work above just uses RNN-LSTMs as a classifier. In this paper, we investigate and design very deep ConvNets for learning action features from skeleton sequences and classify them into classes. Five residual networks [12] were designed and experimented. Our results show state-of-the-art performance on the MSR Action 3D dataset [16] and Kinect Activity Recognition Dataset (KARD) [8].

## 3 Proposed Model

In this section, our proposed model is presented. We first describe a data transformation module which allows us to encode skeleton sequences into 3D-arrays and store them in color images. We then review the ResNet model and propose five different ResNet architectures to solve the task of human skeleton based action recognition.

### 3.1 Data transformation

Currently, the real-time skeleton estimation algorithms have been integrated into depth cameras. This technology allows to extract correctly the position of the joints in the body. E.g., the latest version of the Kinect can help researchers to get more anatomically correct positions for crisp interactions (Figure 1).

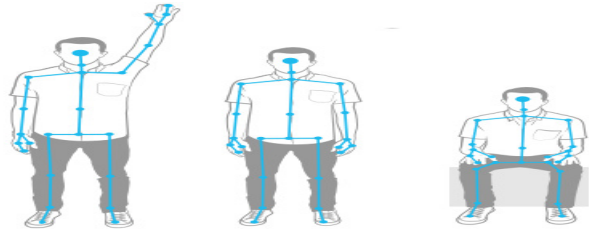

Figure 1. The position of the joints extracted by Kinect [21].

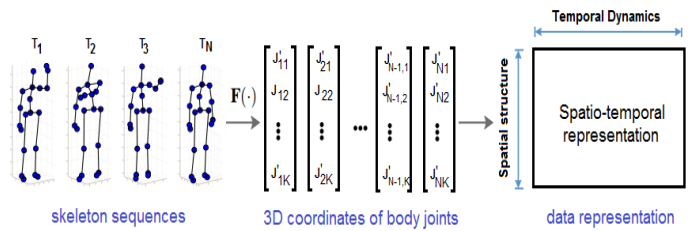

Figure 2. Illustration of the data transformation process. $N$ denotes the number of frames in each sequence. $K$ denotes the number of joints in each frame.

The skeleton data is captured in frames, each frame contains the 3D coordinates of skeletal joints. We transform all the 3D coordinates $(x_i, y_i, z_i)$ of each frame into a new space by normalizing these coordinates by the transformation function $\mathbf{F}(\cdot)$:

$$(x'_i, y'_i, z'_i) = \mathbf{F}(x_i, y_i, z_i)$$
$$x'_i = 255 \times \frac{(x_i - \min\{\mathcal{C}\})}{\max\{\mathcal{C}\} - \min\{\mathcal{C}\}}$$
$$y'_i = 255 \times \frac{(y_i - \min\{\mathcal{C}\})}{\max\{\mathcal{C}\} - \min\{\mathcal{C}\}}$$
$$z'_i = 255 \times \frac{(z_i - \min\{\mathcal{C}\})}{\max\{\mathcal{C}\} - \min\{\mathcal{C}\}}$$

where $(x'_i, y'_i, z'_i)$ are coordinates in the new space; $\max\{\mathcal{C}\}$ and $\min\{\mathcal{C}\}$ are the maximum and minimum values of all coordinates, respectively. Then we stack all normalized frames in order of time $[T_1, T_2, ..., T_N]$ to represent the whole action sequence and store them in RGB color space. All these images are then resized to $40 \times 40$ pixels. Figure 2 illustrates this transformation process. By this way, we converted the skeleton information to 3D tensors which will then be fed into deep learning framework as the input data. Naturally, human body is structured by four limbs and a trunk. Simple actions can be performed through the movement of a limb; more complex actions come from the movements of a group of limbs or the whole body. Inspired by this idea, Du *et al.* [6] proposed an effective representation for skeleton sequences by dividing each skeleton frame into five parts, including two arms $(P_1, P_2)$, two legs $(P_4, P_5)$, and one trunk $(P_3)$ (Figure 3). To have a better representation, we rearrange the pixels in RGB images obtained above in the following order: $P_1 \rightarrow P_2 \rightarrow P_3 \rightarrow P_4 \rightarrow P_5$. Figure 4 shows some examples of images representation using input sequences from MSR Action 3D dataset [16].

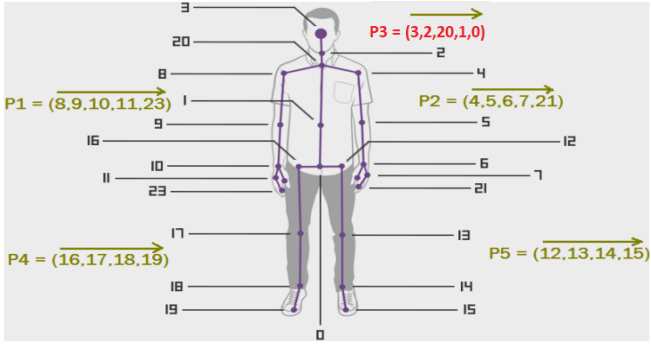

Figure 3. Map of joints in each part from $P_1$ to $P_5$.

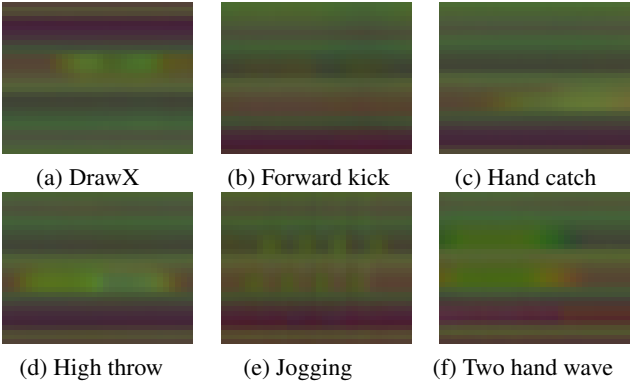

(a) DrawX  (b) Forward kick  (c) Hand catch

(d) High throw  (e) Jogging  (f) Two hand wave

Figure 4. Output of the transformation module obtained from some samples on the MSR Action 3D dataset [16].

## 3.2 Deep residual networks for skeleton based human action recognition

In this section, we first review ResNets and its effectiveness in different recognition tasks. We then propose the use of ResNets for recognizing human action from skeleton sequences.

### 3.2.1 Residual learning

Very deep neural networks demonstrate to have a high performance on many visual-related tasks. However, they are more difficult to train. ResNet [12] is a powerful solution to solve this problem. ResNets allow to make the network training process faster while attaining more accuracy compared to their equivalent models. ResNet is a state-of-the-art ConvNet model which won the 1st place on the ILSVRC 2015 challenge. This model has also achieved the best results on many other tasks related to detection, localization and segmentation. A simple difference between ResNets and traditional ConvNets is that ResNets provide a clear path for gradients to back propagate to early layers in the network. This technique helps the learning process faster. A deep ResNet is constructed from multiple basic blocks that are serially connected to each other. There a shortcut connection parallel between input and output of each block and it gets added to the output of block (Figure 5;left). In other words, a layer of a traditional neural network learns to calculate a mapping function $y = \mathcal{F}(x)$. A ResNet layer approximately calculate $y = \mathcal{F}(x) + id(x)$ where $id(x)$ is a identity function: $id(x) = x$. ResNets use the convolutional layers with $3 \times 3$ filters. Batch normalization [15] and ReLU [22] are applied after each convolution (Figure 5;right). Experimental results on very large datasets for image classification showed that the use of the shortcut connections in ResNet architecture makes the network more accurate and faster.

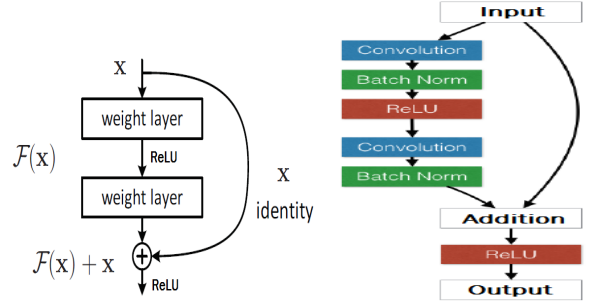

Figure 5. A building block of ResNet.

### 3.2.2 Network design

Motivated by the recent success of ResNets [12], we apply this model to the task of human action recognition in video using skeleton data. We suggest different configurations of ResNet with 20, 32, 44, 56, and 110 layers, denoted by ResNet-20, ResNet-32, ResNet-44, ResNet-56, ResNet-110, respectively. All models are designed for accepting images with size $32 \times 32$ and trained from scratch. The last fully-connected layer of the network represents the action class scores and its size can be changed corresponding to the number of action classes.

## 4 Experiment

### 4.1 Dataset and experimental protocol

#### 4.1.1 MSR Action 3D dataset

We conduct the first experiments on MSR Action 3D dataset [16], a public benchmarking dataset used by many authors for evaluating action recognition algorithms. It contains 20 different actions. Each action is performed by 10 subjects for three times. There are 567 skeleton sequences in total, however 10 sequences are not valid because the skeletons were either missing or wrong, so we conduct our test on 557 sequences. We followed the same experimental protocol as many other authors. More specifically, the whole data is divided into three subsets called **AS1**, **AS2**, **AS3** (Table 1). For each subset, five actors are selected for training and the rest for testing.

#### 4.1.2 Kinect Activity Recognition Dataset (KARD)

The KARD dataset was collected by Gaglio *et al.* [8]. It contains 18 actions, performed by 10 subjects and each subject repeated each action three times for creating a number of 540 sequences in total. KARD is composed of RGB, depth and skeleton frames. Each skeleton frame contains 15 joints. The authors also proposed an evaluation protocol on this data in which the whole dataset is divided into three subsets as shown in Table 2. For each subset, three experiments have been proposed. **Experiment A** uses one-third of the dataset for training and the rest for testing. Meanwhile, **Experiment B** uses two-third of the dataset for training and one-third for testing. Finally, **Experiment C** uses a half of the dataset for training and the rest for testing.

| AS1 | AS2 | AS3 |
|---|---|---|
| *Horizontal arm wave* | *High arm wave* | *High throw* |
| *Hammer* | *Hand catch* | *Forward kick* |
| *Forward punch* | *Draw* x | *Side kick* |
| *High throw* | *Draw tick* | *Jogging* |
| *Hand clap* | *Draw circle* | *Tennis swing* |
| *Bend* | *Two hand wave* | *Tennis serve* |
| *Tennis serve* | *Forward kick* | *Golf swing* |
| *Pickup* & *Throw* | *Side-boxing* | *Pickup* & *Throw* |

Table 1. Three subsets of the MSR Action 3D dataset.

| Action Set 1 | Action Set 2 | Action Set 3 |
|---|---|---|
| *Horizontal arm wave* | *High arm wave* | *Draw tick* |
| *Two-hand wave* | *Side kick* | *Drink* |
| *Bend* | *Catch cap* | *Sit down* |
| *Phone call* | *Draw tick* | *Phone call* |
| *Stand up* | *Hand clap* | *Take umbrella* |
| *Forward kick* | *Forward kick* | *Toss paper* |
| *Draw X* | *Bend* | *High throw* |
| *Walk* | *Sit down* | *Horiz. arm wave* |

Table 2. List of actions in each subset of the KARD dataset.

### 4.2 Data augmentation

Very deep neural networks require a lot of data to train. Unfortunately, we have only 557 skeleton sequences on MSR Action 3D dataset and 540 sequences on KARD dataset. Thus, to prevent overfitting, data augmentation has been applied. We used random cropping, flip horizontally and vertically techniques to generate more data and add them into training set. More specifically, 8 × cropping has been applied on 40 × 40 images to create 32 × 32 images. Then, their horizontally and vertically flipped images are also created. In addition, color effect has been also applied. By this way, we have enough data for each action class and ensure that our ResNet models work well.

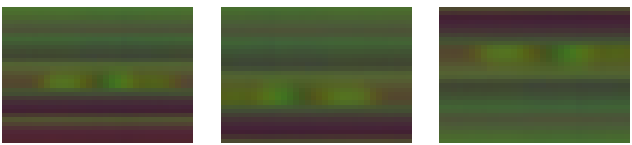

(a) Original image (b) Horizontal flipping (c) Vertical flipping

Figure 6. Data augmentation applied on MSR Action 3D dataset.

| Model | AS1 | AS2 | AS3 | Aver. |
|---|---|---|---|---|
| ResNet-20 | **99.40** | **99.00** | **100.0** | **99.47** |
| ResNet-32 | 99.50 | 98.70 | 99.70 | 99.30 |
| ResNet-44 | 99.60 | 98.20 | 99.80 | 99.20 |
| ResNet-56 | 99.20 | 97.30 | 99.60 | 98.70 |
| ResNet-110 | 99.20 | 98.00 | 99.90 | 99.37 |

Table 3. Test accuracies (%) of our proposed models on AS1, AS2, and AS3 subsets.

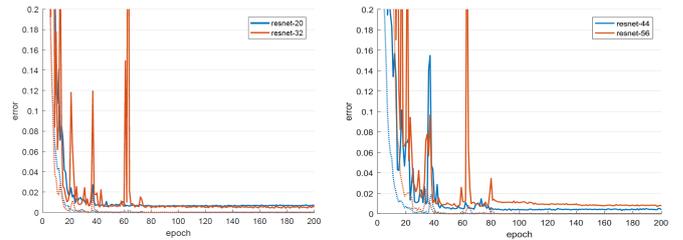

(a) ResNet-20 and ResNet-32  (b) Resnet-44 and ResNet-56

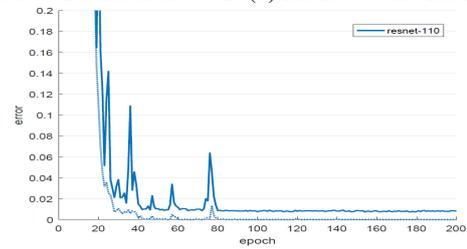

(a) ResNet-110

Figure 7. Learning curves on **AS1**. Dashed lines denote training errors, bold lines denote testing errors.

## 5 Experimental Results

We used MatConvNet [27] to implement our deep learning models. MatConvNet is an open source computing framework for ConvNets. It provides many pre-defined neural network layers, supports efficient computation on GPUs and allows to train state-of-the-art ConvNets. The proposed models are trained and tested on a computer using Geforce GTX 1080 Ti GPU with 11GB RAM.

### 5.1 MSR Action 3D dataset

Experimental results on MSR Action 3D dataset are shown in Table 2. Some learning curves of all the models on **AS1** are shown in Figure 7. We achieved the best classification accuracy with ResNet-20 model. More specifically, classification accuracies are **99.4%** on **AS1**, **99%** on **AS2**, and **100%** on **AS3**. We obtained a total average accuracy of **99.47%**. This result shows that our proposed model outperforms many previous works (Table 5). In addition, our study found that the learning behavior of ResNet depend on the size of the dataset. Specifically, ResNet-20 got better results than ResNet-32, ResNet-44, ResNet-56, and ResNet-110.

|  | Activity Set 1 | | |
| --- | --- | --- | --- |
| **Model** | **Exp. A** | **Exp. B** | **Exp. C** |
| Resnet-20 | 100 | 100 | 100 |
| ResNet-32 | 100 | 100 | 100 |
| ResNet-44 | 100 | 100 | 99.9 |
| ResNet-56 | 100 | 100 | 99.9 |
| ResNet-110 | 99.7 | 100 | 100 |
|  | Activity Set 2 | | |
| **Model** | **Exp. A** | **Exp. B** | **Exp. C** |
| ResNet-20 | 100 | 100 | 100 |
| ResNet-32 | 100 | 100 | 99.9 |
| ResNet-44 | 100 | 100 | 100 |
| ResNet-56 | 100 | 100 | 100 |
| ResNet-110 | 99.9 | 100 | 100 |
|  | Activity Set 3 | | |
| **Model** | **Exp. A** | **Exp. B** | **Exp. C** |
| ResNet-20 | 99.8 | 100 | 99.8 |
| ResNet-32 | 99.8 | 99.9 | 99.8 |
| ResNet-44 | 99.0 | 99.7 | 99.7 |
| ResNet-56 | 99.4 | 99.9 | 99.8 |
| ResNet-110 | 99.1 | 100 | 99.7 |

Table 4. Test accuracies (%) of our proposed models on the KARD dataset.

| **Method** | **AS1** | **AS2** | **AS3** | **Aver.** |
| --- | --- | --- | --- | --- |
| Li *et al.* [16] | 72.90 | 71.90 | 79.20 | 74.67 |
| Vieira *et al.* [31] | 84.70 | 81.30 | 88.40 | 84.80 |
| Xia *et al.* [33] | 87.98 | 85.48 | 63.46 | 78.97 |
| Chaaraoui *et al.* [1] | 92.38 | 86.61 | 96.40 | 91.80 |
| Chen *et al.* [3] | 96.20 | 83.20 | 92.00 | 90.47 |
| Luo *et al.* [19] | 97.20 | 95.50 | 99.10 | 97.26 |
| Gowayyed *et al.* [10] | 92.39 | 90.18 | 91.43 | 91.26 |
| Hussein *et al.* [14] | 88.04 | 89.29 | 94.29 | 90.53 |
| Qin *et al.* [23] | 81.00 | 79.00 | 82.00 | 80.66 |
| Liang *et al.* [17] | 73.70 | 81.50 | 81.60 | 78.93 |
| Evangelidis *et al.* [7] | 88.39 | 86.61 | 94.59 | 89.86 |
| Ilias *et al.* [26] | 91.23 | 90.09 | 99.50 | 93.61 |
| Gao *et al.* [9] | 92.00 | 85.00 | 93.00 | 90.00 |
| Vieira *et al.* [30] | 91.70 | 72.20 | 98.60 | 87.50 |
| Chen *et al.* [2] | 98.10 | 92.00 | 94.60 | 94.90 |
| Du *et al.* [5] | 93.33 | 94.64 | 95.50 | 94.49 |
| **Our best model** | **99.40** | **99.00** | **100.00** | **99.47** |

Table 5. Comparing our performance with other approaches on the MSR Action 3D dataset. All methods used the same experimental protocol.

| **Method** | **Exp. A** | **Exp. B** | **Exp. C** |
| --- | --- | --- | --- |
| Gaglio *et al.* [8] | 89.73 | 94.50 | 88.27 |
| Cippitelli *et al.* [4]; P = 7 | 96.03 | 97.80 | 96.37 |
| Cippitelli *et al.* [4]; P = 11 | 96.47 | 98.27 | 96.87 |
| Cippitelli *et al.* [4]; P = 15 | 96.00 | 97.97 | 96.80 |
| Ling *et al.* [18] | 98.90 | 99.60 | 99.43 |
| **Our best model** | **99.87** | **100.0** | **99.93** |

Table 6. Average recognition accuracy (%) of the best proposed model for experiments A, B and C compared to other approaches on the whole KARD dataset using the same experimental protocol.

### 5.2 KARD dataset

We performed a total of 45 experiments on KARD dataset [8]. The experimental results on KARD dataset are shown in Table 4. Table 6 provides an accuracy comparison between our best proposed model and other approaches on the whole KARD dataset. This result confirms that our approach outperformed the previous state-of-the-art on KARD dataset.

## 6 Conclusion and Future Work

In this paper, we proposed a deep residual framework to learn and recognize human action using skeleton sequences. Experimental results on two challenging action datasets demonstrate the power of ResNets in understanding complex human actions. For future research, some interesting improvements in ResNet architecture are under study by many researchers. Therefore, we believe that this idea will be widely applied for human action recognition in the near future.

### Acknowledgements

This work was supported by the Cerema Research Center and Universidad Carlos III de Madrid. Sergio A. Velastin has received funding from the European Unions Seventh Framework Programme for Research, Technological Development and demonstration under grant agreement No 600371, el Ministerio de Economa, Industria y Competitividad (COFUND2013-51509) el Ministerio de Educacin, cultura y Deporte (CEI-15-17) and Banco Santander.